\pdfoutput=1

\documentclass[11pt]{article}
\usepackage[preprint]{acl}

\usepackage{times}
\usepackage{latexsym}

\usepackage[T1]{fontenc}

\usepackage[utf8]{inputenc}

\usepackage{microtype}

\usepackage{inconsolata}

\usepackage{graphicx}
\usepackage{subfigure}
\usepackage{multirow}
\usepackage{multicol}
\usepackage{booktabs}
\usepackage{xcolor}
\usepackage{colortbl}
\definecolor{LightBlue}{rgb}{0.8235,0.9737,0.9882}

\usepackage{amsmath}
\usepackage{verbatim}
\usepackage{makecell}

\usepackage{hyperref} 

\title{Carrot and Stick: Inducing Self-Motivation with Positive \& Negative Feedback}

\author{
 \textbf{Jimin Sohn\textsuperscript{1}},
 \textbf{Jeihee Cho\textsuperscript{2}},
 \textbf{Junyong Lee\textsuperscript{2}},
 \textbf{Songmu Heo\textsuperscript{3}},
 \textbf{Ji-Eun Han\textsuperscript{4}},
 \textbf{David R. Mortensen\textsuperscript{5}},
 \\
 \textsuperscript{1} GIST, South Korea
 \textsuperscript{2} Yonsei University, South Korea
 \\
 \textsuperscript{3} Korea University, South Korea,
 \textsuperscript{4} KT, South Korea,
 \textsuperscript{5} Carnegie Mellon University, USA
\\
 \small{
   \href{mailto:estelle26598@gm.gist.ac.kr}{estelle26598@gm.gist.ac.kr}
 }
}

\begin{document}
\maketitle
\begin{abstract}
Positive thinking is thought to be an important component of self-motivation in various practical fields such as education and the workplace. Previous work, including sentiment transfer and positive reframing, has focused on the positive side of language. However, self-motivation that drives people to reach their goals has not yet been studied from a computational perspective. Moreover, negative feedback has not yet been explored, even though positive and negative feedback are both necessary to grow self-motivation. To facilitate self-motivation, we propose \textbf{CA}rrot and \textbf{STIC}k (\textbf{\textit{CASTIC}}) dataset, consisting of $12,590$ sentences with 5 different strategies for enhancing self-motivation. Our data and code are publicly available at \href{https://github.com/orgs/EMNLP-self-motivation/repositories}{here}.
\end{abstract}

\section{Introduction}\label{sec:intro}
Interest in positive psychological aspects of language has growing in the field of NLP. \citet{ziems-etal-2022-inducing}, \citet{sharma-etal-2023-cognitive}, and \citet{maddela-etal-2023-training} introduce the task of \textit{Positive Reframing}, aiming to shift the negative perspective of a statement into a positive one without altering the original content. \citet{njoo-etal-2023-talkup} proposed a new benchmark analyzing how empowerment is conveyed in language.

Previous research has only focused on reframing negative thoughts into positive ones, ignoring the value of non-positive language.
In this work, we propose a new approach to appropriately utilize negative (and positive) language as feedback, so as to induce \textit{self-motivation} via stimulation.

\begin{figure}[t!]
    \centering
    \includegraphics[width=0.47\textwidth]{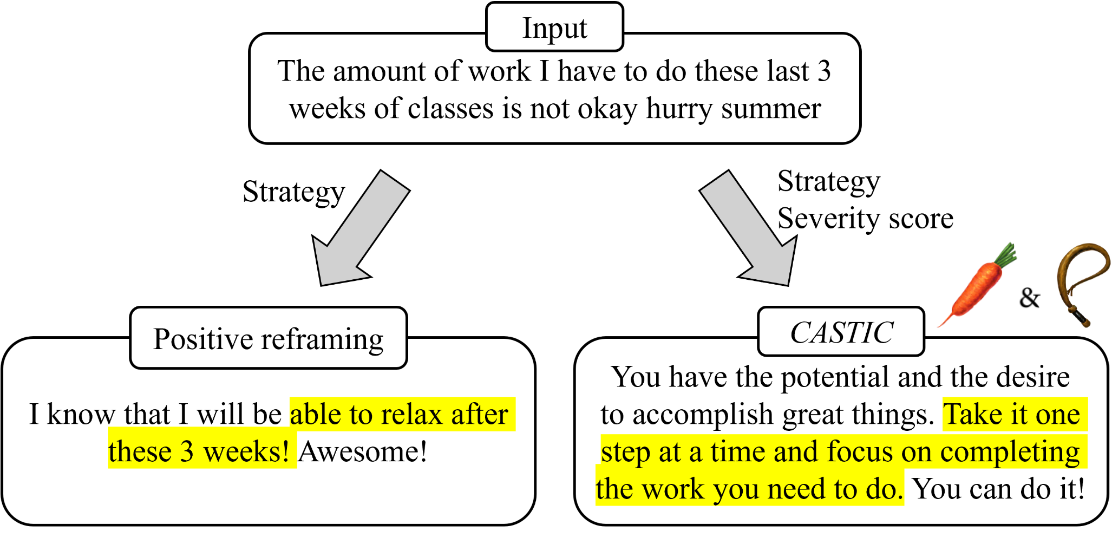}
    \caption{An example of positive reframing~\cite{ziems-etal-2022-inducing} and feedback generated using our \textit{CASTIC} framework.}
    \label{fig:concept_figure}
\end{figure}
Self-motivation is an internal drive that leads a person to take action towards a goal, which is significant in various real-world domains such as education and business. One popular theoretical approach to motivation is \citet{maslow1958dynamic}, proposing that motivation is derived from five basic needs: physiological, safety, belongingness \& love, esteem, and self-actualization, which are hierarchically organized.

Researchers have attempted to enhance the motivation of people by giving feedback relevant to their situations. 
In \citet{kim2019effects}, it was found that when students received negative feedback, they achieved more accurate self-assessment of skills compared to positive feedback, while positive feedback enhanced students' self-efficacy and 
boosted confidence in their ability to achieve goals. Hence, the findings from \citet{kim2019effects} suggest that a balanced use of positive and negative feedback is necessary to optimize self-motivation.
\citet{wisniewski2020power} also concluded that positive feedback was effective in enhancing confidence and motivation while negative feedback 
helped to clearly identify areas of deficiency and motivate improvement. Although negative feedback might seem demotivating, it helped in identifying areas that need improvement, guiding future efforts, and avoiding past mistakes. The analysis of the results for each type of feedback in \citet{wisniewski2020power} also aligns with \citet{kim2019effects}, indicating that for optimal self-motivation, both positive and negative feedback should be used in a balanced manner.

\begin{figure*}[t!]
    \centering
    \includegraphics[width=1.0\textwidth]{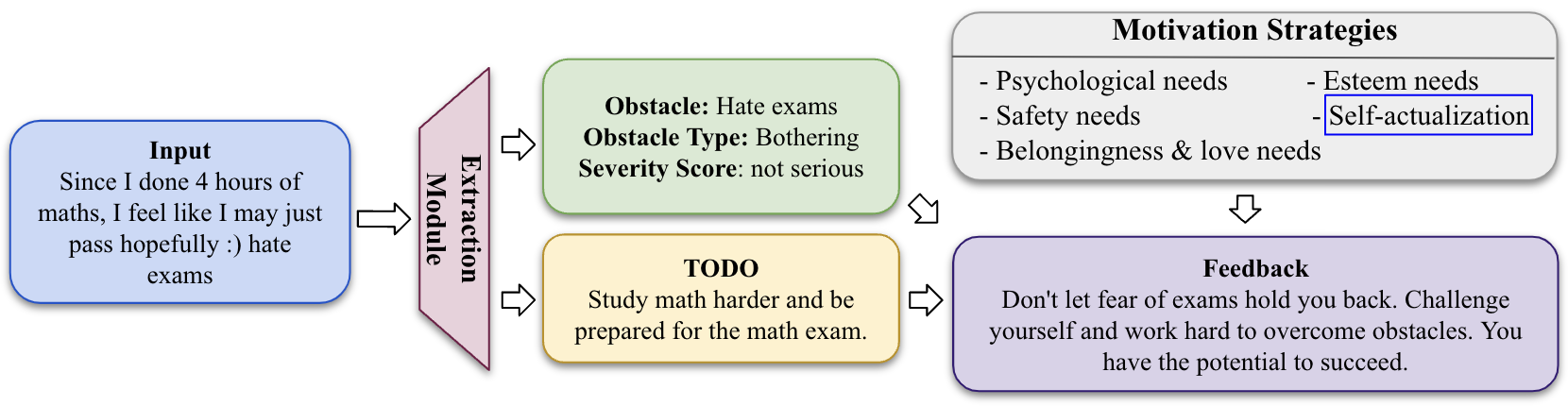}
    \caption{Overall procedure of generating \textit{CASTIC} dataset}
    \label{fig:overall_framework}
\end{figure*}

In this work, we introduce a new benchmark named \textbf{CA}rrot and \textbf{STIC}k (\textbf{\textit{CASTIC}}) meant measure induction of self-motivation.
We address the task by providing both positive and negative feedback. As far as we can determine, dealing with negative aspects of language in the context of motivation is methodologically novel within NLP. The proposed dataset is generated with a three step procedure and evaluated with both quantitative and qualitative approaches.

\section{\textit{CASTIC} Dataset} \label{sec:main}
\begin{table*}[ht!]
\centering
\resizebox{\textwidth}{!}{
\begin{tabular}{lll}
\hline
\textbf{Obstacle Types} & \textbf{Definition \& Example} & \textbf{Severity score}\\
\hline
\multirow{2}{*}{Relationships} & Conflict situations with nearby people & \multirow{2}{*}{not serious} \\
& \emph{fight with friends, mother's nagging} & \\
\midrule
\multirow{2}{*}{Health} & Physical or mental illness & \multirow{2}{*}{serious} \\
& \emph{migraine, stomachache, burn out} & \\
\midrule
\multirow{2}{*}{Fear, overwhelmed} & Anxiety about what will happen in the future or previous failings & \multirow{2}{*}{serious} \\
& \emph{Anxiety about past failures} & \\
\midrule
\multirow{2}{*}{Lack of resource} & A lack of supplies needed for work & \multirow{2}{*}{serious} \\
& \emph{lack of internet connection, lost laptop} & \\
\midrule
\multirow{2}{*}{Annoyance} & Irritation by trivial, annoying situations & \multirow{2}{*}{not serious} \\
& \emph{noisy circumstance} & \\
\midrule
\multirow{2}{*}{Rest, Entertainment} & Lack of motivation due to desire for simple entertainment & \multirow{2}{*}{not serious} \\
& \emph{game, movie, dating} & \\
\midrule
\multirow{2}{*}{etc} & Any situation other than the above & \multirow{2}{*}{serious} \\ 
& \emph{Internet/banking system error, bad weather} & \\
\hline
\end{tabular}
}
\caption{\label{tab:obstacle_types}
The seven types of obstacles blocking someone from reaching their goal and the corresponding severity score. The definition is indicated at the top and a corresponding example is given in italics.
}
\end{table*}

Large Language Models tend to generate sentences with a positive bias \cite{chen2023say}. However, from the perspective of motivation, unconditional positive support is not always what is needed. Stimulating feedback relevant to the person's circumstance is more effective in achieving goals.
Therefore, instead of always giving positive feedback, the model should provide either negative or positive feedback depending on how seriously the obstacle interferes with the task to be done.
To give language models this ability, we propose \textbf{\textit{CASTIC}} dataset that provides appropriate feedback for given sentences. In this section, we present our overall procedure for data generation and provide a taxonomy of the types of obstacles and strategies for giving feedback.

\subsection{Data Collection}
The overall procedure of generating \textit{CASTIC} is provided in Fig.~\ref{fig:overall_framework}.
The prompt for extracting \textit{TODO}, \textit{Obstacle}, and generating the final feedback sentence are provided in Appendix~\ref{appx:datacollection}.

\paragraph{Input sentence} We use input sentences from 
Positive Psychology Frames (POSREF, \cite{ziems-etal-2022-inducing} collected from Twitter with simple keyword \#stressed. We use only the \texttt{original text} column from the dataset.

\paragraph{Obstacle and TODO Extraction Module} We use Orion-14B-Chat~\cite{chen2024orion} with an Apache-2.0 License to generate datasets as it is an open-source large language model (LLM) with outstanding performance in comprehensive evaluations and supporting even extremely long texts \cite{chen2024orion}. The model first extracts \textit{TODO} and \textit{Obstacle} from the input sentence. \textit{TODO} is the goal that people aim to achieve and \textit{Obstacle} is the challenge or obstacle that hinder people from achieving \textit{TODO}. The model gets to respond "None" when there is no specific \textit{TODO} in a given sentence, and the feedback is generated considering only \textit{TODO}.

\paragraph{Obstacle Type and Severity Score} We annotate which of the seven categories the extracted \textit{Obstacle} belongs to. It is worth noting that a sentence can have multiple obstacles and therefore can have multiple types. The \textit{Severity Score} is assigned corresponding to the obstacle type in Table~\ref{tab:obstacle_types}. \textit{Severity Score} means how seriously the \textit{Obstacle} blocks the person from \textit{TODO}. If the sentence has multiple obstacles, the sentence is considered ``not serious'' only when all the obstacle types are considered ``not serious''. Categories and corresponding severity scores were determined according to the criteria by which we manually checked and classified all input data. 

\paragraph{Feedback Generation} To generate feedback inducing self-motivation, we use five motivation strategies referenced from the widely known Motivation Theory's ``five needs'' \cite{maslow1958dynamic}. A detailed explanation of each need is provided in Appendix~\ref{appx:motivation_thoery}. Feedback was created with LLM (Orion-14B-Chat) using the \textit{TODO}, \textit{Obstacle}, and \textit{Severity Score} from the previous step and each of the five needs. The severity score determines whether the feedback is positive or negative, and each of five needs determines which aspects to emphasize to motivate the person. We reviewed each feedback generated by the model.

\subsection{Data Distribution}
We evaluate the statistics of seven obstacle types in our CASTIC dataset in Table \ref{tab:obstacle_number}. As one sample can have multiple obstacle types, the total number does not indicate the number of distinct samples. The statistics of frequently appearing words in the dataset is provided in Appendix~\ref{appx:frequency}.

\begin{table}
\centering
\small
\begin{tabular}{ccc}
\hline
\textbf{Obstacle Types} & \textbf{Train \#} & \textbf{Validation \#}\\
\hline
Relationships & 110 & 80 \\
Health & 1,000 & 305 \\
Fear & 5,165 & 1,205 \\
Lack of resource & 235 & 80 \\
Annoyance & 2,015 & 785 \\
Rest & 20 & 50 \\
etc & 1,455 & 135 \\
\midrule
serious & 7,855 & 1,725 \\
not serious & 2,145 & 915 \\
\midrule
Total & 10,000 & 2,640 \\
\hline
\end{tabular}
\caption{\label{tab:obstacle_number}
Summary statistics of each obstacle type in the CASTIC dataset.}
\end{table}

\section{Self-Motivation Framework} \label{sec:eval}
\begin{table*}[ht!]
\centering
\begin{tabular}{cccccccccc}
\hline
Fine-tune & Model & Param. & R-1 & R-2 & R-L & BLEU & BScore & Sim & PPL \\ 
\hline
\multirow{6}{*}{w/o Fine-tune} & GPT & 116M & 11.79 & 0.47 & 8.35 & 0.08 & 82.20 & 0.121 & - \\
& M2M-100 & 483M & 2.10 & 0.16 & 1.89 & 3.20 & 75.66 & 0.088 & 176.36\\
& T5 & 60M & 0.49 & 0.00 & 0.49 & 1.66 & 84.52 & 0.452 & 295.87 \\
& Falcon & 7B & 10.59 & 0.76 & 7.49 & 0.19 & 82.07 & 0.19 & 106.02 \\
& Mistral & 7B & 12.47 & 1.04 & 8.76 & 0.29 & 82.77 & 0.16 & 202.30\\
& BART-L & 406M & 18.10 & 3.76 & 13.00 & 1.81 & 84.94 & 0.458 & 188.65 \\
\hline
\multirow{6}{*}{w/ Fine-tune} &GPT  & 116M & 27.68 & 6.62 & 20.11 & 3.77 & 88.11 & 0.429 & 69.52\\
& M2M-100 & 483M & 30.12 & 8.76 & 22.44 & 5.99 & 88.74 & 0.437 & 32.84\\
& T5 & 60M & 30.28 & 10.04 & \textbf{23.74} & 5.5 & 88.76 & 0.480 & 25.40 \\
& Falcon & 7B & 29.63 & 8.59 & 20.84 & 4.59 & 88.30 & \textbf{0.522} & 33.49\\
& Mistral & 7B & 27.96 & \textbf{13.19} & 23.66 & 2.64 & 88.63 & 0.496 & 27.61\\
& BART-L & 406M & \textbf{33.93} & 10.04 & 23.59 & \textbf{7.00} & \textbf{88.98} & 0.498 & \textbf{24.11} \\
\hline
\end{tabular}
\caption{\label{tab:overall_result}
\textbf{Self-Motivation results} Performance of models with and without fine-tuning on \textbf{\textit{CASTIC}} dataset on ROUGE-1 (R-1), ROUGE-1 (R-2), ROUGE-L (R-L), BLEU, and BERTScore (BScore). Param., Sim, PPL indicates the number of parameters of each model, cosine similarity and perplexity, respectively.}
\end{table*}

\subsection{Task Formulation}
To generate feedback, we extract the \textbf{Obstacle} $O_i$ and \textbf{TODO} $T_i$ from the input sentence. Then, the \textbf{Severity Score} $SS_i$ is assigned based on the \textbf{Obstacle Type} $OT_i$ of the $O_i$. Then, \textbf{Feedback Type} $FT_i$ is labeled as either positive or negative according to the severity score. 
$$ FT_i = \begin{cases} \text{Positive if} \; SS_i=\text{serious}, \\\text{Negative if} \;SS_i=\text{not serious} \end{cases} $$
Based on \textbf{Motivation Strategy} $M_i$, the final \textbf{Feedback} $F_i$ to induce self-motivation is generated.
$$ F_i = \{FT_i, M_i, O_i, T_i\} $$

\subsection{Evaluation}
\subsubsection{Experimental Setup}
\textbf{Dataset} The CASTIC dataset contains 9,990 samples in the train split and 2,600 samples in the validation split. All the data are in English. \\
\textbf{Model} We use BART-L \cite{lewis2019bart}, GPT-2 \cite{radford2019language}, M2M-100 \cite{fan2021beyond}, T5 \cite{raffel2020exploring} model to test the dataset. The number of parameters per model is described in Table \ref{tab:overall_result}.

\subsubsection{Evaluation Metric}
\textbf{Quantitative metric} In various studies, the BLEU \cite{papineni-etal-2002-bleu} and ROUGE \cite{lin2004rouge} metrics are utilized to evaluate the semantic similarity with the ground truth. In this paper, we use BLEU, ROUGE-1, ROUGE-2, ROUGE-L, and BERTScore \cite{zhang2019bertscore} for qualitative results following previous work \cite{ziems-etal-2022-inducing}. Cosine similarity between generated output and ground truth is measured using the sentence transformer all-MiniLM-L6-v2 \cite{reimers-2019-sentence-bert}. Perplexity \cite{bengio2000neural} is measured using GPT-2 \cite{radford2019language} from Hugging Face. It is worth noting that we discard the empty generation samples when measuring the scores.\\
\textbf{Qualitative metric} Following~\citet{chiang2023can}, we use GPT-3.5 \cite{brown2020language} to evaluate the effect of our dataset on inducing self-motivation. The prompt is illustrated in Appendix~\ref{appx:llm_eval}. From the model's generated feedback, we randomly sample 100 sentences and ask GPT-3.5 how motivating (Motivation) and how fluent (Fluency) the feedback is. The rating scale is from 1-5 with 1 being the lowest.

\subsection{Experimental Result}
\paragraph{Overall Result} In Table \ref{tab:overall_result}, we evaluate the result of the experiment. The models can learn each motivation strategy and generate feedback well. We illustrate the example of generated feedback in Table~\ref{tab:bart_example} of Appendix~\ref{appx:qualitative_result}. Overall, BART-L shows the best performance both in zero-shot and fine-tuning experiments.

\begin{table}[t!]
\centering
\resizebox{\columnwidth}{!}{
\begin{tabular}{cccccc}
\hline
\textbf{Strategy} & GPT & GPT-2 & M2M-100 & T5 & BART-L \\
\hline
Physiological Needs & 62.49 & 76.17 & 74.9 & 77.51 & 74.45 \\
Safety Needs & 72.23 & 75.26 & 73.31 & 74.64 & 71.94  \\
Love and Belonging & 72.83 & 76.97 & 76.2 & 76.67 & 65.95 \\
Self-actualization & 67 & 75.32 & 75.87 & 74.95 & 72.32 \\
Esteem Needs & 78.21 & 80.41 & 79.01 & 80.8 & 68.88 \\
\midrule
AVG. & 70.55 & 76.83 & 75.86 & \textbf{76.91} & 70.71  \\
STD. & 6.01 & 2.12 & \textbf{2.09} & 2.48 & 3.32 \\
\hline
\end{tabular}
}
\caption{\label{tab:motivation_result}
F1 score (\%) of motivation strategies classification.
}
\end{table}

\paragraph{LLM Evaluation Result}
We compare the score for models that are fine-tuned on the \textbf{\textit{CASTIC}} dataset and pre-trained models without fine-tuning. In Table \ref{tab:llm_eval_result}, we illustrate the self-motivation feedback resulting from GPT-3.5. The models fine-tuned with our dataset show better performance compared to the others. Specifically, the average motivation score for the fine-tuned models is 2.7, whereas models without fine-tuning achieve an average motivation score of 1.66. Additionally, in terms of fluency, the fine-tuned models attain an average score of 4, outperforming the zero-shot result. The result indicates that the model fine-tuned on our dataset generates fluent outputs.

\begin{table}
\centering
\resizebox{\columnwidth}{!}{
\begin{tabular}{c|cc|cc}
\hline
\multirow{2}{*}{Model}& \multicolumn{2}{c|}{w/ Fine-tune} & \multicolumn{2}{c}{w/o Fine-tune} \\
\cmidrule{2-5}
& Motivation & Fluency & Motivation & Fluency \\ 
\hline
GPT & 2.3 & 3.57 & 1.17 & 2.03 \\
M2M-100 & 2.32 & 3.85 & 1.00 & 1.00 \\
BART-L & 2.54 & 4.34 & 1.82 & 3.06 \\
T5 & 3.03 & 4.18 & 2.12 & 2.47 \\
Mistral & 3.05 & 3.82 & 1.82 & 2.67 \\
Falcon & 2.99 & 4.35 & 2.03 & 2.10 \\
\hline
\end{tabular}
}
\caption{\label{tab:llm_eval_result}
\textbf{LLM Evaluation results} The average rate of generated feedbacks on with and without fine-tuned models in the terms of how motivating and fluent the feedback is.}
\end{table}

\paragraph{Motivation Strategy Classification} In Table \ref{tab:motivation_result}, we evaluate F1 score per 5 motivation strategies. We add one additional linear layer that outputs 5 classes corresponding to each motivation type.
The result shows that each model can learn and distinguish the characteristics between motivation strategies with over 70\% F1 score for every strategy.

\section*{Limitation} \label{sef:limit}
We acknowledge that the severity score, which is determined based on the severity of the obstacle, can be subjective. However, just as people previously judged the negative and positive levels of words and annotated them to obtain negative scores in the field of Sentimental analysis, we present the standards by creating our own dataset and annotating it.

More significantly, we did not test the output of the trained models on human participants to determine, empirically, whether they induced greater levels of self-motivation.

\section*{Ethics Statement}
In this work, we used POSREF~\cite{ziems-etal-2022-inducing} which is a publicly available dataset. The creators of POSREF have already considered ethical issues when creating the dataset, but we additionally have manually checked every input sentence and filtered out inappropriate ones. We didn't find any obvious ethical concerns, such as violent or offensive content. We used the dataset consistent with the intended use. We used LLM in the process of creating and validating the dataset and we performed the verification of the output ourselves, meaning there were no issues with human annotators.


\bibliography{acl_latex}

\appendix

\section{Appendix}
\label{sec:appendix}

\subsection{Implementation Detail}\label{appx:impl_detail}
\textbf{Motivation Result} We train BART-L and M2M-100 with learning rate 1e-4, and batch size 32 for 5 epochs with a maximum sequence length of 128. For GPT and T5, we use learning rate 3e-5, and a weight decay of 0.01 for 5 epochs. For training and inference, we use a NVIDIA H100 80GB GPU. We run the experiment once for all the models. \\
\textbf{Motivation Strategy Classification} We train each model with learning rate 1e-4 and batch size 32 for 2 epochs with output class number as 5. For training and inference, we use a NVIDIA H100 80GB GPU. We run the experiment once for all the models.

\subsection{Maslow's Motivation Theory}\label{appx:motivation_thoery}
In this section, we provide detailed explanation of Maslow's Motivation Theory \cite{maslow1958dynamic}'s five needs.\\
\textbf{Physiological Needs} are desires to maintain a constant, normal state such as with respect to homeostasis, hormones, and nutrients. We applied this strategy for desiring rest, sleep, food, water, air and homeostasis.\\
\textbf{Safety Needs} are the desires for a safe environment such as financial and health security, stability in employment, protection from accidents, and a safe environment. The need derives from human's nature or preferring a safe, orderly, predictable, organized world and disfavoring unexpected, unmanageable, or other dangerous things.\\
\textbf{Love \& Belonging Needs} are desires for love and affection from friends, lovers or spouses. Humans strive to have affectionate relations with people.\\
\textbf{Esteem Needs} are desires for stable, firmly based, high evaluation of themselves. Humans want to feel real respect from themselves and others for their capacities and achievements.\\
\textbf{Self-actualization} refers to the desire for self-fulfillment in actualizing a person's full potential. Humans want to become everything that one is capable of becoming.

\subsection{Prompt for Dataset Generation}
\label{appx:datacollection}
\subsubsection{Obstacle and TODO Extraction}\label{appx:extract}
\begin{figure*}[t!]
    \centering
    \includegraphics[width=0.8\textwidth]{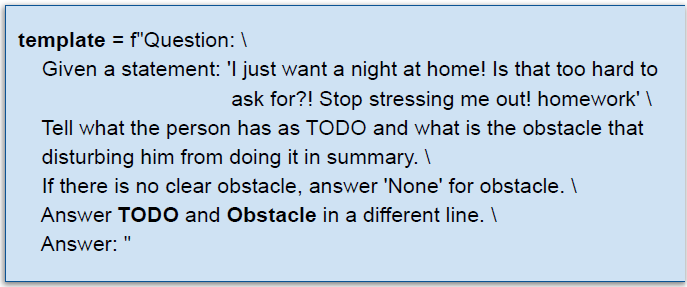}
    \caption{Prompt for Obstacle and TODO extraction}
    \label{fig:obstacle_todo_prompt}
\end{figure*}

We extract \textit{Obstacle} and \textit{TODO} of the negative input sentence with Orion-14B-Chat using the prompt in Figure~\ref{fig:obstacle_todo_prompt}.

\subsubsection{Generating Feedback}\label{appx:feedback_generation}
We generate feedback inducing self-motivation with Orion-14B-Chat using prompt in Figure \ref{fig:obstacle_todo_prompt}. The type of feedback, positive/negative determined by severity score was applied in the feedback.
\begin{figure*}[t!]
    \centering
    \includegraphics[width=0.8\textwidth]{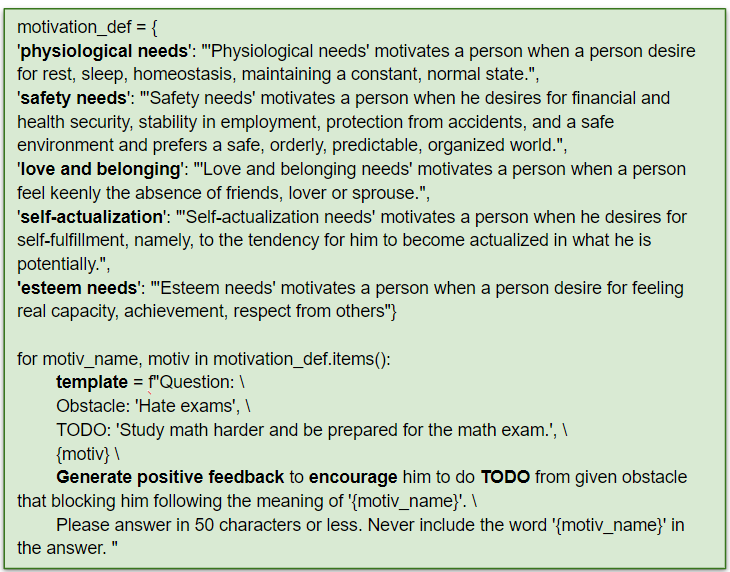}
    \caption{Prompt for Generating Feedback}
    \label{fig:generating_feedback_prompt}
\end{figure*}

\subsection{Prompt for LLM Evaluation}\label{appx:llm_eval}
\begin{figure*}
    \centering
    \includegraphics[width=0.8\textwidth]{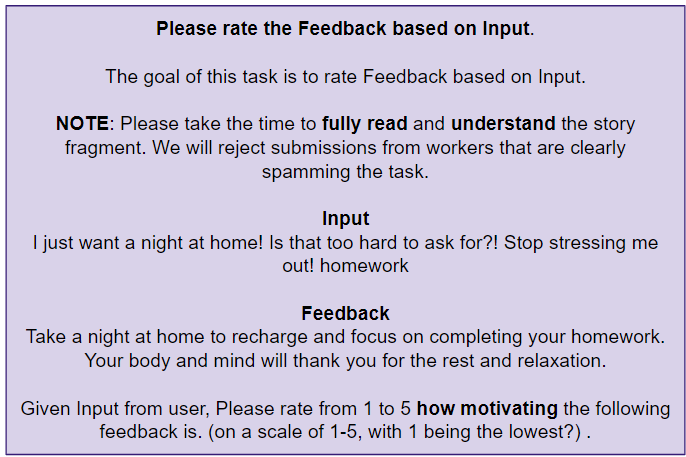}
    \caption{Prompt for evaluating motivation with LLM}
    \label{fig:llm_motivation_prompt}
\end{figure*}

\begin{figure*}
    \centering
    \includegraphics[width=0.8\textwidth]{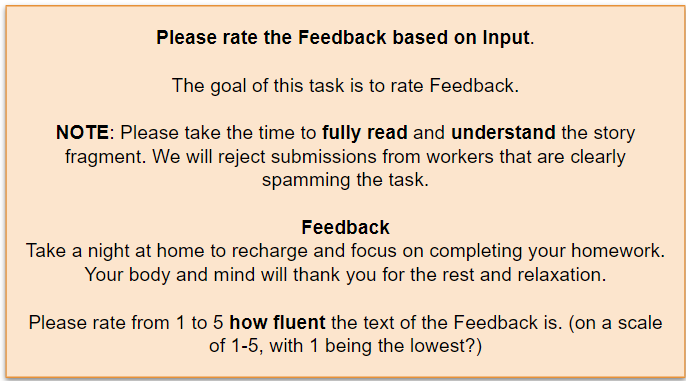}
    \caption{Prompt for evaluating fluency with LLM}
    \label{fig:llm_fluency_prompt}
\end{figure*}

\textbf{Motivation} We evaluate how motivating the feedback generated by model trained in our dataset is with Chat-GPT using prompt in Figure \ref{fig:llm_motivation_prompt} following \cite{chiang2023can}. \\
\textbf{Fluency} We evaluated how fluent the generated feedback by model trained in our dataset is using prompt in Figure \ref{fig:llm_fluency_prompt}.

\subsection{Samples of CASTIC dataset}\label{appx:dataset_example}
In Table \ref{tab:dataset_example}, we illustrate the example of dataset.

\begin{figure*}[t!]
    \centering
    \subfigure[Input]{
        \includegraphics[width=0.4\textwidth]{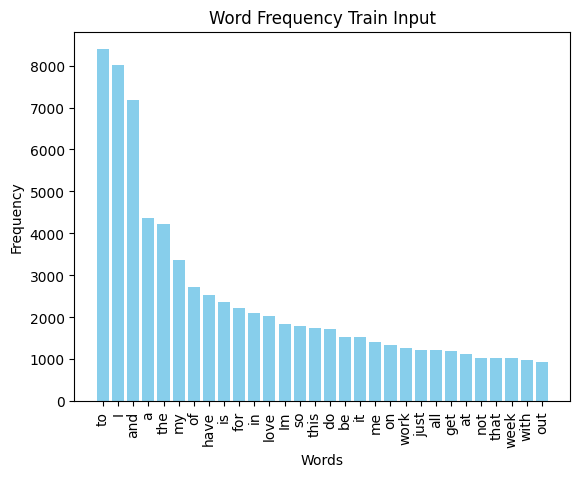}
        \label{fig:traininputfreq}}
    \subfigure[Feedback]{
        \includegraphics[width=0.4\textwidth]{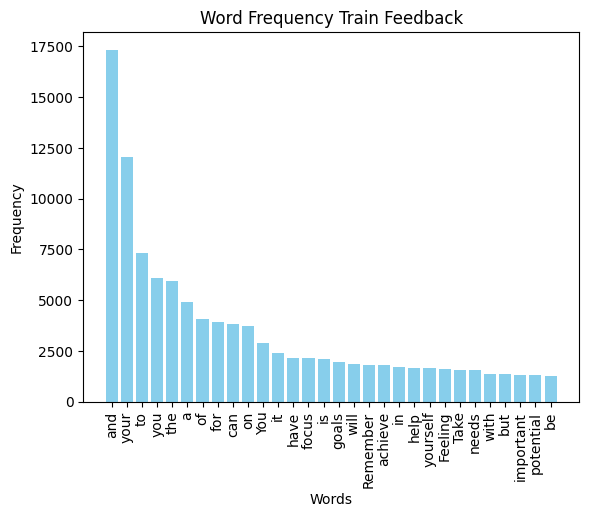}
        \label{fig:trainfeedfreq}}
    \caption{Word frequency analysis for train of \textbf{\textit{CASTIC}}}
    \label{fig:trainwordfreq}
\end{figure*}

\begin{figure*}[t!]
    \centering
    \subfigure[Input]{
        \includegraphics[width=0.4\textwidth]{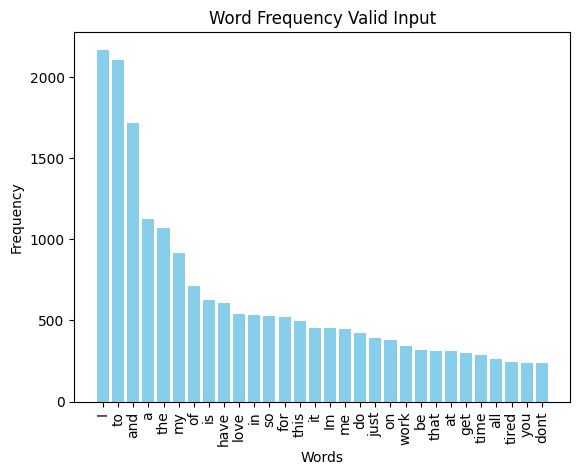}
        \label{fig:validinputfreq}}
    \subfigure[Feedback]{
        \includegraphics[width=0.4\textwidth]{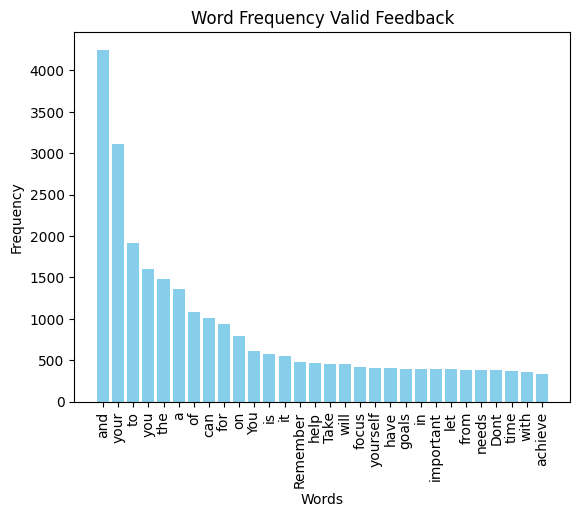}
        \label{fig:validfeedfreq}}
    \caption{Word frequency analysis for validaiton of \textbf{\textit{CASTIC}}}
    \label{fig:validwordfreq}
\end{figure*}

\subsection{Word frequency analysis}\label{appx:frequency}
The word frequency figure illustrates the distribution of terms within our CASTIC dataset, providing key insights into the linguistic patterns and content characteristics of the corpus. As depicted in Figure \ref{fig:traininputfreq}, \ref{fig:trainfeedfreq}, \ref{fig:validinputfreq}, \ref{fig:validfeedfreq}, the x-axis represents individual words ordered by their frequency, while the y-axis shows the occurrence count without prompt words. The figure reveals several prominent trends: notably, common words such as `goal', `love', and `focus' dominate the corpus, reflecting the dataset's focus on emotional motivation. These findings underscore the dataset's thematic concentration and provide a foundation for understanding its linguistic structure and contextual relevance within the field of natural language processing. Moreover, comparative analyses with benchmark datasets reveal distinctive word usage patterns unique to our dataset, highlighting its potential contributions to advancing research in automated language understanding and generation.

\begin{table*}
\centering
\resizebox{\textwidth}{!}{
\begin{tabular}{c|c|c|c|c|c|c}
\hline
\textbf{Input} & \textbf{Obstacle} & \textbf{TODO} & \textbf{Obstacle Type} & \textbf{Severity score} & \textbf{Motivation strategy} & \textbf{Feedback} \\
\hline
\multirow{15}{*}{\begin{tabular}{@{}c@{}}
Still home from school\\
sick gonna pay big time when i\\
get back should be doing homework\\
all weekend on the plane :$($\\
\end{tabular}}
& \multirow{15}{*}{Sick} &
\multirow{15}{*}{\begin{tabular}{@{}c@{}}
Take some rest and do \\
homework all weekend \\
on the plane. \\
\end{tabular}} & \multirow{15}{*}{Health} & \multirow{15}{*}{serious} & physiological need &
\begin{tabular}{@{}c@{}}
Take care of yourself\\
by resting and doing \\
your homework on the plane.\\
Your body will thank you for it! \\
\end{tabular} \\
\cline{6-7}
& & & & & safety needs &
\begin{tabular}{@{}c@{}}
Take care of yourself, prioritize \\
your health and well-being. Rest \\
and study on the plane will help \\
you feel recharged and ready  \\
to tackle any obstacles that come your way. \\
\end{tabular} \\
\cline{6-7}
& & & & & love and belonging &
\begin{tabular}{@{}c@{}}
Take care of yourself and focus \\
on your studies. Your well-being is\\
important for your future relationships. \\
\end{tabular} \\
\cline{6-7}
& & & & & self-actualization &
\begin{tabular}{@{}c@{}}
Take care of yourself and prioritize\\
your well-being. Rest and study will\\
help you achieve your goals\\
and grow as a person. \\
\end{tabular} \\
\cline{6-7}
& & & & & esteem needs &
\begin{tabular}{@{}c@{}}
Rest and study will boost \\
your health and increase your \\
confidence, giving you the chance \\
to show your true potential and achieve success. \\
\end{tabular} \\
\hline
\end{tabular}
}
\caption{Examples of feedback in the dataset.}
\label{tab:dataset_example}
\end{table*}

\subsection{Qualitative result}\label{appx:qualitative_result}
In table, we examine one original text for each of the motivation strategies in our dataset along with the BART model. 

\begin{table*}
\centering
\small
\begin{tabular}{c|c|c}
\hline
\textbf{Input} & \textbf{Motivation strategy} & \textbf{Feedback} \\
\hline
\multirow{23}{*}{
\begin{tabular}{@{}c@{}}
I already hate using computers, \\
so having to apply for a job \\
online is not for me. \\
\end{tabular}
}
& physiological need &
\begin{tabular}{@{}c@{}}
You have the capability and determination to succeed \\
in applying for a job online. Take it one step at \\
a time and focus on what you need to accomplish. \\
\emph{Think about how much easier it will be to apply} \\
\emph{for a job online without having to leave your} \\
\emph{house! You can take care of business and still} \\ 
\emph{maintain a constant, normal state.} \\
\end{tabular} \\
\cline{2-3}
& safety needs &
\begin{tabular}{@{}c@{}}
You have the capability and determination to succeed \\
in applying for a job online. Take it one step at \\
a time and focus on what you can control. You can do it! \\
\emph{Don't let fear of computers hold you back. Apply} \\
\emph{for that job online and take control of your future.} \\
\end{tabular} \\
\cline{2-3}
& love and belonging &
\begin{tabular}{@{}c@{}}
Feeling lonely? Apply for a job online and make new \\
connections. You'll find love and belonging, which \\
are important for your overall well-being. \\
\emph{Feeling lonely? Finding a job online could connect} \\ 
\emph{you with new friends and a sense of community.} \\
\end{tabular} \\
\cline{2-3}
& self-actualization &
\begin{tabular}{@{}c@{}}
You have the potential and the desire to succeed. \\
Let's focus on applying for a job online \\
and fulfilling your potential. \\
\emph{Don't let fear of computers hold you back.} \\ 
\emph{Take control and apply for that job} \\
\emph{online. You have the potential to succeed} \\
\emph{and this is your chance to show it.} \\
\end{tabular} \\
\cline{2-3}
& esteem needs &
\begin{tabular}{@{}c@{}}
You have the capability and determination to succeed. \\
Let's focus on applying for the job \\
online and show your skills. \\
\emph{Feeling capable and respected is important, but} \\
\emph{hating computers won't get you far. It's time to} \\
\emph{overcome your fear and apply for that job online.} \\
\end{tabular} \\
\hline
\end{tabular}
\caption{\label{tab:bart_example} Examples of generated feedback of BART in a controlled setting. Prediction feedback corresponds to each motivation strategy is at the first row and the ground truth follows in italic.}
\end{table*}

\end{document}